\RequirePackage[l2tabu, orthodox]{nag} %

\documentclass{article}

\PassOptionsToPackage{numbers, compress}{natbib}

\usepackage[final]{bdu_2024}

\usepackage{amsmath} %
\usepackage{amssymb} %
\usepackage{amsfonts} %
\usepackage{amsopn} %
\DeclareMathOperator{\diag}{diag}
\usepackage{mathrsfs} %
\usepackage{euscript} %
\usepackage{mathtools}
\let\hbar\hbarorig   %

\usepackage{algorithm} %
\usepackage{algpseudocode} %

\usepackage{caption}
\usepackage{subcaption}

\algnewcommand\algorithmicinput{\textbf{Input:}}
\algnewcommand\algorithmicoutput{\textbf{Output:}}
\algnewcommand\algorithmicnote{\textbf{Note:}}
\algnewcommand\Input{\item[\algorithmicinput]}%
\algnewcommand\Output{\item[\algorithmicoutput]}%
\algnewcommand\Note{\item[\algorithmicnote]}%

\usepackage{threeparttable} %
\usepackage{booktabs}       %

\usepackage{graphicx} %

\usepackage[hidelinks]{hyperref}       %
\usepackage{url}            %
\usepackage{xcolor}         %

\usepackage{microtype} %
\usepackage[utf8]{inputenc} %
\usepackage[T1]{fontenc}    %
\usepackage[strings]{underscore} %
\usepackage[strict=true]{csquotes} %
\usepackage{nicefrac}       %
\usepackage{cleveref}       %

\makeatletter
\renewcommand{\paragraph}{%
  \@startsection{paragraph}{4}{\z@}%
  {0.1\baselineskip}  %
  {-0.15em}            %
  {\normalfont\normalsize\bfseries}%
}
\makeatother

\usepackage{enumitem}

\title{Gaussian Process Thompson Sampling via Rootfinding}

\author{%
  Taiwo A. Adebiyi \\ %
  University of Houston\\
 \texttt{taadebiyi2@uh.edu} \\
  \And
  Bach Do \\
  University of Houston\\
  \texttt{bdo3@uh.edu} \\
  \And
  Ruda Zhang \\
  University of Houston\\
  \texttt{rudaz@uh.edu} \\
}

\begin{document}

\maketitle

\begin{abstract}
Thompson sampling (TS) is a simple, effective stochastic policy in Bayesian decision making.
It samples the posterior belief about the reward profile
and optimizes the sample to obtain a candidate decision.
In continuous optimization, the posterior of the objective function is often a Gaussian process (GP),
whose sample paths have numerous local optima, making their global optimization challenging.
In this work, we introduce an efficient global optimization strategy for GP-TS
that carefully selects starting points for gradient-based multi-start optimizers.
It identifies all local optima of the prior sample via univariate global rootfinding,
and optimizes the posterior sample using a differentiable, decoupled representation.
We demonstrate remarkable improvement in the global optimization of GP posterior samples,
especially in high dimensions.
This leads to dramatic improvements in the overall performance of Bayesian optimization
using GP-TS acquisition functions,
surprisingly outperforming alternatives like GP-UCB and EI.
\end{abstract}

\vspace{-1em}
\section{Introduction}

Bayesian optimization (BO) is a highly successful approach to the global optimization of
expensive-to-evaluate black-box functions \cite{Garnett2023}.
Effectively, there are two nested iterations in BO:
the outer-loop seeks to optimize the objective function $f(\mathbf{x})$;
and the inner-loop seeks to optimize the acquisition function $\alpha(\mathbf{x})$ at each stage.
The premise of BO is that the inner-loop optimization can be solved accurately and efficiently,
so that the outer-loop proceeds informatively with a negligible added cost.
In fact, the convergence guarantees of many BO strategies
assume \textit{exact} global optimization of the acquisition function \cite{Srinivas2010, Bull2011, Russo2014}.
However, this is more challenging than commonly assumed \cite{Wilson2018}.

Gaussian process Thompson sampling (GP-TS) \cite{Kandasamy2018} uses posterior samples directly as acquisition functions.
It has strong theoretical guarantees \cite{Russo2014},
scales to high dimensions \cite{Mutny2018},
and can be easily parallelized \cite{Shah2015, HernandezLobato2017, Kandasamy2018}.
It is also used in computing information-theoretic acquisition functions
such as predictive entropy search \cite{HernandezLobato2014} and max-value entropy search \cite{WangZ2017}.
But posterior sample functions are notoriously difficult to optimize due to their complexity.

We present an efficient strategy that globally optimizes GP-TS acquisition functions
by judiciously selecting starting points for gradient-based multi-start optimizers.
It exploits the separability of multivariate GP priors and the decomposition of GP posterior samples per ``Matheron's rule'' \cite{Wilson2020}.
The former allows for the identification of all local minima of a GP prior sample using a robust rootfinding algorithm;
the latter links the prior sample and the data to a posterior sample, relates their critical points, and facilitates the selection of starting points.

\section{Spectral Representation of Gaussian Processes}

\paragraph{Gaussian Processes.}
Given a training dataset $\mathcal{D}=\left\{ (\mathbf{X},\mathbf{y})\right\}=\{(\mathbf{x}^i,y^i)\}_{i=1}^N$,
where $\textbf{x}^i$ is an input location and $y^i$ is the corresponding observation.
Define an observation model $y(\mathbf{x})=f(\mathbf{x}) + \epsilon$,
where $f(\mathbf{x})$ is an unknown objective function and
$\epsilon \overset{\text{iid}}{\sim}  \mathcal{N}(0,\sigma_\text{n}^2)$
is independent and identically distributed (iid) zero-mean Gaussian noise.

A GP assumes that any finite subset of function values has a joint Gaussian distribution \cite{Rasmussen2006}.
This assumption is encoded in the GP prior $f(\mathbf{x}) \sim \mathcal{GP} \left(m(\mathbf{x}),\kappa(\mathbf{x},\mathbf{x}')\right)$,
where $m(\mathbf{x})$ is the mean function and $\kappa(\mathbf{x},\mathbf{x}')$ a positive definite covariance function.
Conditioning the GP prior on the data provides a posterior that is also a GP.
Assuming $m(\mathbf{x}) = 0$, the GP posterior can be written as
$f(\mathbf{x}_\star)|\mathcal{D} \sim \mathcal{GP} \left(\widehat{m}(\mathbf{x}_\star), \widehat{\kappa}(\mathbf{x}_\star,\mathbf{x}'_\star)\right)$.
Here the posterior mean
$\widehat{m}(\mathbf{x}_\star)= \mathbf{k}^\intercal(\mathbf{x}_\star,\mathbf{X}) \mathbf{C}^{-1} \mathbf{y}$
and the posterior covariance $\widehat{\kappa}(\mathbf{x}_\star, \mathbf{x}'_\star) = \kappa(\mathbf{x}_\star, \mathbf{x}'_\star) - \mathbf{k}^\intercal(\mathbf{x}_\star,\mathbf{X}) \mathbf{C}^{-1} \mathbf{k}(\mathbf{x}'_\star,\mathbf{X})$,
where $\mathbf{C} = \mathbf{K} + \sigma_\text{n}^2 \mathbf{I}$ with $ \mathbf{K}_{ij} = \kappa(\mathbf{x}^i,\mathbf{x}^j)$ ($i,j \in \{1\dots,N\}$), and $\mathbf{k}(\mathbf{x}_\star,\mathbf{X}) = \left[\kappa(\mathbf{x}_\star,\mathbf{x}^1),\dots,\kappa(\mathbf{x}_\star,\mathbf{x}^N)\right]^\intercal$.

\paragraph{Spectral Representation of Gaussian Processes.}
Per Mercer's theorem on probability spaces \cite{Rasmussen2006}, any positive definite covariance function that is essentially bounded with respect to some probability measure $\mu$ on the domain $\mathcal{X}$ has a spectral representation $\kappa(\mathbf{x}, \mathbf{x}') = \sum_{k=0}^\infty \lambda_k \phi_k(\mathbf{x}) \phi_k(\mathbf{x}')$,
where $(\lambda_k, \phi_k(\mathbf{x}))$ is a pair of eigenvalue and eigenfunction of the kernel integral operator.
A sample function from the GP prior can thus be written as
$f(\mathbf{x}) = \sum_{i=0}^{\infty} w_k \sqrt{\lambda_k} \phi_k(\mathbf{x})$,
where  $w_k \overset{\text{iid}}{\sim} \mathcal{N}(0,1)$.

\paragraph{Decoupled Representation of GP Posteriors.} 
Given a GP prior sample $f(\mathbf{x})$, we can determine a posterior sample $\widetilde{f}(\mathbf{x})$
using a decoupled representation that updates the prior sample according to Matheron's rule \cite{Wilson2020}.
For observations contaminated by iid Gaussian noise, we have
$\widetilde{f}(\mathbf{x}_\star) = f(\mathbf{x}_\star) + \mathbf{k}^\intercal (\mathbf{x}_\star,\mathbf{X}) \mathbf{C}^{-1} \left( \mathbf{y} - \mathbf{f} - \boldsymbol{\epsilon} \right)$,
where $\mathbf{f} = \left[ f(\mathbf{x}^1),\dots,f(\mathbf{x}^N) \right]^\intercal$ and $\boldsymbol{\epsilon} \overset{\text{iid}}{\sim} \mathcal{N}_N(\mathbf{0}, \sigma_\text{n}^2 \mathbf{I}_N)$.

\section{Global Optimization to Thompson Sampling Acquisition Functions}

\paragraph{Assumptions.}
We use a GP prior with a separable covariance function.
We require that the univariate components of this covariance function either have known spectral representations per Mercer's theorem,
or have known spectral densities per Bochner's theorem with an effective discretization (see e.g., \cite{Solin2020, Mutny2018}),
and that the corresponding samples are continuously differentiable.
The squared exponential (SE) covariance function meets these requirements and is of our focus in the following.
We further assume that the objective function is defined on a hypercube $\mathcal{X} = \prod_{i=1}^d [\underline{x}_i, \overline{x}_i]$.

\paragraph{Spectrum of SE Covariance Function.}
Consider the univariate SE covariance function $\kappa(x, x'; l) = \exp\left(-\frac{1}{2}(x - x')^2/l^2\right)$, where $l$ is the characteristic length scale.
Per Mercer's theorem, it has a spectral representation $\kappa(x, x') = \sum_{k=0}^{\infty} \lambda_k \phi_k(x) \phi_k(x')$. 
For a Gaussian measure $\mu = \mathcal{N}(0, \sigma^2)$ over the real line, let $a = (2 \sigma^{2})^{-1}$, $b = (2 l)^{-1}$, $c = \sqrt{a^2 + 4 a b}$, and
$A = \frac{1}{2} a + b + \frac{1}{2} c$.
For $k \in \mathbb{N}$, the $k$th eigenvalue is $\lambda_k = \sqrt{a/A} \left( b/A\right)^k$ and the corresponding eigenfunction is
$\phi_k(x) = \left( \pi c/a \right)^{1/4} \psi_k (\sqrt{c} x) \exp\left( \frac{1}{2} a x^2 \right)$,
where $\psi_k (x) = \left( \pi^{1/2} 2^k k! \right)^{-1/2} H_k(x) \exp\left( -\frac{1}{2}x^2 \right)$ and $H_k(x) = (-1)^k \exp(x^2) \frac{d^k}{dx^k} \exp(-x^2)$ the $k$th-order Hermite polynomial (see e.g., \cite{ZhuHY1998} Sec. 4).
In a general case with $k = (k_1,\dots,k_d)$, $\lambda_k = \prod_{j=1}^{d} \lambda_{k_j}$ and $\phi_k (\mathbf{x}) = \prod_{j=1}^{d} \phi_{k_j}(x_j)$, where  $\lambda_k$ and $\phi_k (\mathbf{x})$ are the eigenvalues and eigenfunctions of the $d$-variate covariance function $\kappa(\mathbf{x}, \mathbf{x}')$, respectively.

\paragraph{Prior Sample Functions.}
The separability of the covariance function and the known spectral representations of the component covariance functions
allow us to accurately approximate the prior sample as
$f(\mathbf{x}) \approx  \prod_{i=0}^{d} \sum_{k=1}^{N_i-1} w_{i,k} \sqrt{\lambda_{i,k}}  \phi(x_i)$.
Here $N_i$ is selected for each variate such that $\lambda_{i,N_i-1}/\lambda_{i,1} \leq \eta_i$,
where $\eta_i$ is sufficiently small, e.g., $\eta_i = 10^{-16}$.

\paragraph{Properties of Posterior Sample Functions.}
With the decoupled representation of GP posteriors, each posterior sample has the form of
$\widetilde{f}(\mathbf{x}) = f(\mathbf{x}) + b(\mathbf{x})$.
Here, the prior sample $f(\mathbf{x}) = \prod_{i=1}^d f_i(x_i)$ is fast-varying and separable,
enabling the use of efficient univariate root-finding algorithms \cite{Trefethen2019} to identify all its critical points.
The data adjustment $b(\mathbf{x}) = \sum_{j=1}^N v_j \kappa(\mathbf{x}, \mathbf{x}^j)$, where $v_j \in \mathbb{R}$,
is a weighted sum of canonical basis functions.
While not separable, it is smoother, has much fewer critical points than the prior sample, and has limited effect away from data points.

\paragraph{Critical Points of Multivariate Separable Functions.}
The critical points of the multivariate separable prior sample $f(\mathbf{x}) = \prod_{i=1}^{d} f_i(x_i)$
are exactly the critical points of its univariate components $f_i(x_i)$, arbitrarily combined, except for when $f(\mathbf{x}) = 0$.
As a result, we can find all the relevant critical points of the prior sample function $f(\mathbf{x})$ by solving a global rootfinding problem for the derivative of each of its univariate components:
$f_i'(x_i) = 0$, $i \in \{1, \cdots, d\}$.
We also add the upper and lower bounds of each variable $x_i$ to the set of critical points of $f_i$,
as they can define the extrema of $f(\mathbf{x})$ on the bounded domain.
Let $\{\zeta_{i,j}\}_{j=1}^{r_i}$ represent the set of critical points of $f_i(x_i)$.

\paragraph{Local Minima of Multivariate Separable Functions.}
Given the univariate critical points,
identifying a local minimum of a multivariate separable prior sample $f(\mathbf{x})$ is straightforward.
Let $\mathbf{x} = (\zeta_{i,s(i)})_{i=1}^d$ %
be an arbitrary combination of the univariate critical points.
If $\mathbf{x}$ is an interior point, it is a local minimum if $\nabla^2f(\mathbf{x}) \succ 0$,
which has the form $\nabla^2f(\mathbf{x}) = \diag\{ \prod_{i \neq j} f_j(x_j) f''_i(x_i) \}_{i= 1}^d$.
If $\mathbf{x}$ is a boundary point, the criterion is slightly modified.
Without enumerating all combinations, best subsets of the local minima $\mathcal{S}_{\min}$ can be efficiently identified as follows:
(1) filter the critical points for local extrema, exploiting the structures of $\nabla^2f(\mathbf{x})$ and $\nabla f(\mathbf{x})$;
(2) select the local extrema with the largest $|f|$, using a max heap data structure; and
(3) local minima have negative $f$ values.

\begin{figure}[t]
    \centering
    \includegraphics[width=\textwidth]{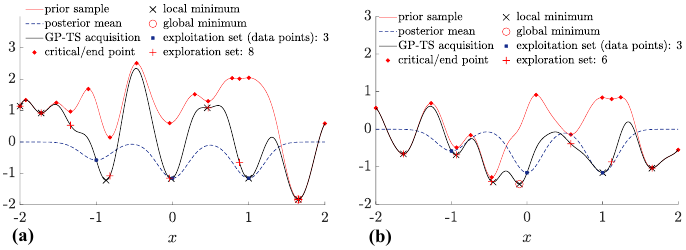}
    \caption{Illustration of exploration and exploitation sets for global optimization of GP-TS acquisition functions. (a) When the global minimum of the GP-TS acquisition function lies outside the interpolation region, it is typically identified by starting the gradient-based optimizer at a local minimum of the prior sample. (b) When the global minimum is within the interpolation region, it can be found by starting the gradient-based optimizer at either a data point or a local minimum of the prior sample. } 
     \label{fig:GlobalOpt}
\end{figure}

\paragraph{Global Optimization to Thompson Sampling.}
To globally optimize a GP-TS acquisition function in each BO iteration,
we use two sets of starting points for a gradient-based multi-start optimizer,
namely exploitation $\mathcal{S}_\text{p}$ and exploration $\mathcal{S}_\text{e}$.
Here $\mathcal{S}_\text{p}$ contains the data points, while $\mathcal{S}_\text{e}$ is either $\mathcal{S}_{\min}$,
or a subset of $\mathcal{S}_{\min}$ when the number of its members is too large, e.g., $> 1000$.
\Cref{fig:GlobalOpt} illustrates the exploration and exploitation sets for global optimization of two GP-TS acquisition functions.

Our optimization strategy is motivated by a few observations.
When the prior sample $f$ is added to the smoother landscape of the data adjustment $b$,
each local minimum of $f$ will be located nearby a local minimum of the posterior sample $\widetilde{f}$.
Searching from data points can discover good local minima of $\widetilde{f}$ in the vicinity of the data set,
which can pickup some local minima not readily discovered by the local minima of $f$.
This is especially true if $f$ is relatively flat near a data point.
Starting from both $\mathcal{S}_\text{e}$ and $\mathcal{S}_\text{p}$
can thus give sufficient coverage of the best local minima of $\widetilde{f}$,
leading to efficient global optimization of GP-TS acquisition functions.

\begin{figure}[t]
    \centering
    \includegraphics[width=\textwidth]{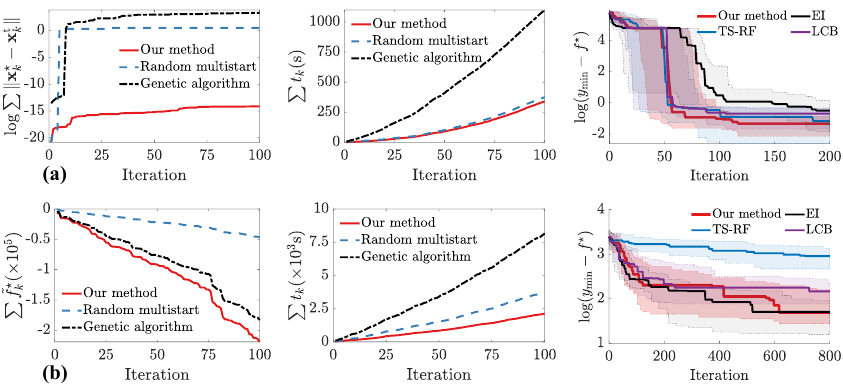}
     \caption{Optimization results for (a) 2d Schwefel and (b) 10d Levy functions. \textit{Top-left}: Cumulative distances between new candidate solutions $\mathbf{x}_k^\star$ to the true global minimums $\mathbf{x}_k^\text{t}$ of the GP-TS acquisition functions $\widetilde{f}(\mathbf{x})$ for 2d Schwefel function. \textit{Bottom-left}: Cumulative optimized values $\widetilde{f}_k^\star$ for 10d Levy function. \textit{Middle}: Cumulative run time $t_k$ required for optimizing $\widetilde{f}(\mathbf{x})$. \textit{Right}: Histories of medians and interquartile ranges of solutions from 20 runs of our method, TS-RF, EI, and LCB.} 
     \label{fig:optResults}
\end{figure}

\section{Experiments}

\paragraph{Inner-Loop Optimization.}
We minimize the 2d Schwefel and 10d Levy functions \cite{Surjanovic2013} to assess the quality of the inner-loop solutions recommended by our proposed method.
We start with $10d$ data points, normalize the input data to $[-1,1]^d$, and standardize the output data. We set $\sigma = 1$ and $\sigma_\text{n}=10^{-6}$.
For comparison, we optimize the same GP-TS acquisition functions using a gradient-based multi-start optimizer with random starting points (i.e., random multi-start) and a genetic algorithm. The number of starting points for the random multi-start and the population size of the genetic algorithm are equal to the number of starting points of our method. The same stopping criteria are used for the three algorithms.

\Cref{fig:optResults} compares the inner-loop solutions and CPU times for inner-loop optimization by our method, random multi-start, and genetic algorithm.
For both problems, our method outperforms the random multi-start and genetic algorithm in terms of the inner-loop solution quality and the run time required for optimizing GP-TS acquisition functions. This verifies the importance of judicious selection of starting points for optimizing GP-TS acquisition functions in both low and high dimensions.

\paragraph{Outer-Loop Optimization.}
 We compare the outer-loop solutions by our method with those by other BO methods, including TS with random Fourier features (TS-RF) \cite{Rahimi2007,HernandezLobato2014}, expected improvement (EI) \cite{Jones1998}, and lower confidence bound (LCB)---the version of GP-UCB \cite{Srinivas2010} for minimization.
The inner-loop optimization for TS-RF, EI, and LCB is performed via a gradient-based multi-start optimizer with random starting points.
The number of starting points and the termination criteria for this optimizer are the same as those for our method.
In each BO iteration, we record the log simple regret $\log(y_{\min} - f^\star)$,
where $y_{\min}$ is the best observation up to that iteration and $f^\star$ is the true minimum of the objective function.

\Cref{fig:optResults} shows the medians and interquartile ranges of solutions from 20 runs of each BO method for the test functions.
On the 2d Schwefel function, our method can achieve better objective function values than all other considered methods.
It also provides a competitive result in optimizing the 10d Levy function.
Across the two examples, EI and LCB tend to perform well in the initial iterations,
while our method shows fast improvement in later iterations,
highlighting the exploratory nature and delayed reward of the GP-TS policy.
Considering robustness to the objective function,
GP-TS (perhaps surprisingly) outperforms EI and LCB, when optimized using our method.
  
\section{Summary}
We propose a method to optimize GP-TS acquisition functions globally. 
The method relies on local minima of prior samples obtained from a univariate rootfinding algorithm, the data points, and a gradient-based multi-start optimizer with carefully selected starting points.
Its effectiveness is supported by the prevalent use of separable covariance functions in BO,
where the univariate covariance components is expressed in terms of their spectral representations.
The optimization results show that the proposed method offers higher-quality solutions to optimizing GP-TS acquisition functions
in both low- and high-dimensional settings, compared to a random multi-start and a genetic algorithm.
It also shows dramatic improvements in outer-loop optimization.
\clearpage

\bibliography{bduTSroots}

\begin{thebibliography}{19}
\providecommand{\natexlab}[1]{#1}
\providecommand{\url}[1]{\texttt{#1}}
\expandafter\ifx\csname urlstyle\endcsname\relax
  \providecommand{\doi}[1]{doi: #1}\else
  \providecommand{\doi}{doi: \begingroup \urlstyle{rm}\Url}\fi

\bibitem[Garnett(2023)]{Garnett2023}
Roman Garnett.
\newblock \emph{Bayesian Optimization}.
\newblock Cambridge University Press, Cambridge, 2023.
\newblock \doi{10.1017/9781108348973}.

\bibitem[Srinivas et~al.(2010)Srinivas, Krause, Kakade, and
  Seeger]{Srinivas2010}
Niranjan Srinivas, Andreas Krause, Sham~M. Kakade, and Matthias Seeger.
\newblock {Gaussian process optimization in the bandit setting: No regret and
  experimental design}.
\newblock In \emph{Proceedings of the 27th International Conference on Machine
  Learning}, volume~13, pages 1015--1022, 2010.
\newblock URL \url{https://icml.cc/Conferences/2010/papers/422.pdf}.

\bibitem[Bull(2011)]{Bull2011}
Adam~D. Bull.
\newblock Convergence rates of efficient global optimization algorithms.
\newblock \emph{Journal of Machine Learning Research}, 12\penalty0
  (88):\penalty0 2879--2904, 2011.
\newblock URL \url{http://jmlr.org/papers/v12/bull11a.html}.

\bibitem[Russo and Van~Roy(2014)]{Russo2014}
Daniel Russo and Benjamin Van~Roy.
\newblock Learning to optimize via posterior sampling.
\newblock \emph{Mathematics of Operations Research}, 39\penalty0 (4):\penalty0
  1221--1243, 2014.
\newblock \doi{10.1287/moor.2014.0650}.

\bibitem[Wilson et~al.(2018)Wilson, Hutter, and Deisenroth]{Wilson2018}
James Wilson, Frank Hutter, and Marc Deisenroth.
\newblock Maximizing acquisition functions for {B}ayesian optimization.
\newblock In \emph{Advances in Neural Information Processing Systems},
  volume~31, pages 9884--9895, 2018.
\newblock URL
  \url{https://proceedings.neurips.cc/paper/2018/hash/498f2c21688f6451d9f5fd09d53edda7-Abstract.html}.

\bibitem[Kandasamy et~al.(2018)Kandasamy, Krishnamurthy, Schneider, and
  Poczos]{Kandasamy2018}
Kirthevasan Kandasamy, Akshay Krishnamurthy, Jeff Schneider, and Barnabas
  Poczos.
\newblock Parallelised {B}ayesian optimisation via {T}hompson sampling.
\newblock In \emph{Proceedings of the Twenty-First International Conference on
  Artificial Intelligence and Statistics}, volume~84, pages 133--142, 2018.
\newblock URL \url{https://proceedings.mlr.press/v84/kandasamy18a.html}.

\bibitem[Mutny and Krause(2018)]{Mutny2018}
Mojmir Mutny and Andreas Krause.
\newblock Efficient high dimensional {B}ayesian optimization with additivity
  and quadrature {F}ourier features.
\newblock In \emph{Advances in Neural Information Processing Systems},
  volume~31, pages 9005--9016, 2018.
\newblock URL
  \url{https://proceedings.neurips.cc/paper/2018/hash/4e5046fc8d6a97d18a5f54beaed54dea-Abstract.html}.

\bibitem[Shah and Ghahramani(2015)]{Shah2015}
Amar Shah and Zoubin Ghahramani.
\newblock Parallel predictive entropy search for batch global optimization of
  expensive objective functions.
\newblock In \emph{Advances in Neural Information Processing Systems},
  volume~28, pages 3330--3338, 2015.
\newblock URL
  \url{https://proceedings.neurips.cc/paper_files/paper/2015/file/57c0531e13f40b91b3b0f1a30b529a1d-Paper.pdf}.

\bibitem[Hern{\'a}ndez-Lobato et~al.(2017)Hern{\'a}ndez-Lobato, Requeima,
  Pyzer-Knapp, and Aspuru-Guzik]{HernandezLobato2017}
Jos{\'e}~Miguel Hern{\'a}ndez-Lobato, James Requeima, Edward~O. Pyzer-Knapp,
  and Al{\'a}n Aspuru-Guzik.
\newblock Parallel and distributed {T}hompson sampling for large-scale
  accelerated exploration of chemical space.
\newblock In \emph{Proceedings of the 34th International Conference on Machine
  Learning}, volume~70, pages 1470--1479, 2017.
\newblock URL \url{https://proceedings.mlr.press/v70/hernandez-lobato17a.html}.

\bibitem[Hern{\'{a}}ndez-Lobato et~al.(2014)Hern{\'{a}}ndez-Lobato, Hoffman,
  and Ghahramani]{HernandezLobato2014}
Jos{\'{e}}~Miguel Hern{\'{a}}ndez-Lobato, Matthew~W. Hoffman, and Zoubin
  Ghahramani.
\newblock Predictive entropy search for efficient global optimization of
  black-box functions.
\newblock In \emph{Advances in Neural Information Processing Systems},
  volume~27, pages 918--926, 2014.
\newblock URL
  \url{https://proceedings.neurips.cc/paper_files/paper/2014/hash/069d3bb002acd8d7dd095917f9efe4cb-Abstract.html}.

\bibitem[Wang and Jegelka(2017)]{WangZ2017}
Zi~Wang and Stefanie Jegelka.
\newblock Max-value entropy search for efficient {B}ayesian optimization.
\newblock In \emph{Proceedings of the 34th International Conference on Machine
  Learning}, volume~70, pages 3627--3635, 2017.
\newblock URL \url{https://proceedings.mlr.press/v70/wang17e.html}.

\bibitem[Wilson et~al.(2020)Wilson, Borovitskiy, Terenin, Mostowsky, and
  Deisenroth]{Wilson2020}
James Wilson, Viacheslav Borovitskiy, Alexander Terenin, Peter Mostowsky, and
  Marc Deisenroth.
\newblock Efficiently sampling functions from {G}aussian process posteriors.
\newblock In \emph{Proceedings of the 37th International Conference on Machine
  Learning}, volume 119, pages 10292--10302, 2020.
\newblock URL \url{https://proceedings.mlr.press/v119/wilson20a.html}.

\bibitem[Rasmussen and Williams(2006)]{Rasmussen2006}
Carl~Edward Rasmussen and Christopher K.~I. Williams.
\newblock \emph{Gaussian processes for machine learning}.
\newblock The MIT Press, 2006.
\newblock ISBN 9780521872508.
\newblock \doi{10.7551/mitpress/3206.001.0001}.

\bibitem[Solin and Särkkä(2020)]{Solin2020}
Arno Solin and Simo Särkkä.
\newblock Hilbert space methods for reduced-rank {G}aussian process regression.
\newblock \emph{Statistics and Computing}, 30\penalty0 (2):\penalty0 419--446,
  2020.
\newblock \doi{10.1007/s11222-019-09886-w}.

\bibitem[Zhu et~al.(1998)Zhu, Williams, Rohwer, and Morciniec]{ZhuHY1998}
Huaiyu Zhu, Christopher K.~I. Williams, Richard Rohwer, and Michal Morciniec.
\newblock Gaussian regression and optimal finite dimensional linear models.
\newblock In \emph{Neural Networks and Machine Learning}, 1998.
\newblock URL \url{https://publications.aston.ac.uk/id/eprint/38366/}.

\bibitem[Trefethen(2019)]{Trefethen2019}
Lloyd~N. Trefethen.
\newblock \emph{Approximation Theory and Approximation Practice, Extended
  Edition}, volume 164.
\newblock Society for Industrial and Applied Mathematics, Philadelphia, PA,
  2019.
\newblock ISBN 978-1-61197-593-2.
\newblock \doi{10.1137/1.9781611975949}.

\bibitem[Surjanovic and Bingham(2013)]{Surjanovic2013}
Sonja Surjanovic and Derek Bingham.
\newblock Virtual library of simulation experiments: Test functions and
  datasets, 2013.
\newblock URL \url{http://www.sfu.ca/~ssurjano}.

\bibitem[Rahimi and Recht(2007)]{Rahimi2007}
Ali Rahimi and Benjamin Recht.
\newblock Random features for large-scale kernel machines.
\newblock In \emph{Advances in Neural Information Processing Systems},
  volume~20, pages 1177--1184, 2007.
\newblock URL
  \url{https://proceedings.neurips.cc/paper_files/paper/2007/file/013a006f03dbc5392effeb8f18fda755-Paper.pdf}.

\bibitem[Jones et~al.(1998)Jones, Schonlau, and Welch]{Jones1998}
Donald~R. Jones, Matthias Schonlau, and William~J. Welch.
\newblock Efficient global optimization of expensive black-box functions.
\newblock \emph{Journal of Global Optimization}, 13\penalty0 (4):\penalty0
  455--492, 1998.
\newblock \doi{10.1023/A:1008306431147}.

\end{thebibliography}

\end{document}